\documentclass[12pt]{article}

\usepackage[letterpaper,top=1in, bottom=1in, left=1in, right=1in]{geometry}

\usepackage{mathtools}
\usepackage{algorithm}
\usepackage{algpseudocode}
\usepackage{graphicx}
\usepackage{setspace}    
\usepackage{lineno}
\usepackage{caption}
\usepackage{cite}    

\usepackage[labelfont=bf]{caption}
\DeclareCaptionLabelSeparator{bar}{\space\textbar\space}
\begin{document}

\large
\noindent \textbf{Neuromorphic Hebbian learning with magnetic tunnel junction\\ synapses}
\vspace{5mm}
\normalsize

\noindent Peng Zhou$^{1}$, Alexander J. Edwards$^{1}$, Frederick B. Mancoff$^{2}$, Sanjeev Aggarwal$^{2,*}$, Stephen K. Heinrich-Barna$^{1,3}$, Joseph S. Friedman$^{1,*}$ 

\vspace{5mm}

\noindent$^{1}$ Department of Electrical and Computer Engineering, The University of Texas at Dallas, Richardson, TX, USA

\noindent$^{2}$ Everspin Technologies, Inc., Chandler, AZ, USA

\noindent$^{3}$ Texas Instruments Inc., Dallas, TX, USA

\noindent$^{*}$ Corresponding Authors: sanjeev.aggarwal@everspin.com; joseph.friedman@utdallas.edu

\newpage
\textbf{Abstract}

Neuromorphic computing aims to mimic both the function and structure of biological neural networks to provide artificial intelligence with extreme efficiency. Conventional approaches store synaptic weights in non-volatile memory devices with analog resistance states, permitting in-memory computation of neural network operations while avoiding the costs associated with transferring synaptic weights from a memory array. However, the use of analog resistance states for storing weights in neuromorphic systems is impeded by stochastic writing, weights drifting over time through stochastic processes, and limited endurance that reduces the precision of synapse weights. Here we propose and experimentally demonstrate neuromorphic networks that provide high-accuracy inference thanks to the binary resistance states of magnetic tunnel junctions (MTJs), while leveraging the analog nature of their stochastic spin-transfer torque (STT) switching for unsupervised Hebbian learning. We performed the first experimental demonstration of a neuromorphic network directly implemented with MTJ synapses, for both inference and spike-timing-dependent plasticity learning. We also demonstrated through simulation that the proposed system for unsupervised Hebbian learning with stochastic STT-MTJ synapses can achieve competitive accuracies for MNIST handwritten digit recognition. By appropriately applying neuromorphic principles through hardware-aware design, the proposed STT-MTJ neuromorphic learning networks provide a pathway toward artificial intelligence hardware that learns autonomously with extreme efficiency.

\newpage


Spiking neural networks (SNNs) can be used in energy-efficient neuromorphic computing systems that process information from spikes or pulses emitted by artificial neurons \cite{merolla2014million,painkras2013spinnaker,davies2018loihi}. The spiking signals flow through artificial synapses, in which the synaptic weights are stored in an energy-free manner via the resistances of non-volatile memory devices. The SNN can be trained through unsupervised Hebbian learning rules, enabling energy-efficient biomimetic learning that does not require labeled data \cite{hebb1949organization,dayan1999unsupervised,bi2001synaptic,dan2004spike}. However, it is challenging to precisely write and store non-volatile resistance states in conventional analog memory devices (\textit{e.g.}, memristors/RRAM, phase change memory (PCM), and ferroelectric devices) \cite{mochida20184m,cai2019fully,wan2022compute,sebastian2018tutorial,ambrogio2018equivalent,dutta2020supervised,khaddam2021hermes,moon2023parallel,luo2021design}, as they suffer from stochastic writing, weights drifting over time through stochastic processes, and limited endurance \cite{chen2013endurance, kim2019phase,ambrogio2014statistical,yu2011investigating,wang2020resistive,shao2022efficient,esmanhotto2022experimental}. In concert with the imprecision and variation inevitable when fabricating large memory arrays, the stochasticity inherent to analog memory devices significantly degrades accuracy of the vector-matrix multiplications (VMM) central to neural network computations. This stochasticity has led to the suppression of analog resistance states and the use of only binary states \cite{chen2019cmos,hung2021four,xue201924,xue2021cmos,golonzka2019non,benoist201428nm}.

Spin-transfer torque (STT) magnetic tunnel junctions (MTJs) naturally provide stable binary resistance states that enable more accurate neural network inferences, while their inherent stochastic switching naturally provides the analog behavior necessary for learning. In particular, Querlioz \textit{et al.} \cite{querlioz2012bioinspired, vincent2015spin} proposed a Hebbian learning rule through which the stochasticity intrinsic to the STT-MTJ switching process emulates spike-timing-dependent plasticity (STDP) learning. Querlioz \textit{et al.} further provided an initial exploration of a neuromorphic architecture for learning and inference with STT-MTJ synapses, as well as behavioral simulations demonstrating the viability of this Hebbian learning approach with binary non-volatile memory devices. Preliminary experiments by Goodwill \textit{et al.} \cite{goodwill2022implementation} and Jung \textit{et al.} \cite{jung2022crossbar} have supported the suitability of MTJs for neuromorphic computing, but included neither unsupervised learning nor a direct implementation of VMM with MTJ synapses.

In this work, we therefore propose and experimentally demonstrate the first complete neuromorphic inference and learning system that exploits the stochastic switching of STT-MTJ synapses. We have performed the first experimental demonstration of a neuromorphic network directly implemented with MTJ synapses, successfully matching patterns using a 4x2 MTJ neural network that recognizes 2x2 pixel images. We have also proposed unsupervised Hebbian neuromorphic learning networks with STT-MTJ synapses, and have experimentally demonstrated the ability of this 4x2 MTJ neural network to leverage STT-MTJ stochastic switching for input image clustering. This work concludes with large-scale simulations of unsupervised Hebbian learning of the MNIST handwritten digit dataset, illustrating the ability of the proposed neuromorphic approach to scale to application-relevant dimensions.

\section*{Experimental demonstration of STT-MTJ binary neuromorphic network}

The core component of the proposed neuromorphic learning networks is the MTJ, which consists of a fixed layer, a tunnel barrier, and a free layer. MTJs, which are conventionally used for magnetoresistive random-access memory (MRAM), have two stable non-volatile states described by the magnetization of the free layer relative to that of the fixed layer: a parallel (P) state with high conductance and an anti-parallel (AP) state with low conductance. 

\begin{figure}[t!]
    \centering
    \includegraphics[width=1\textwidth]{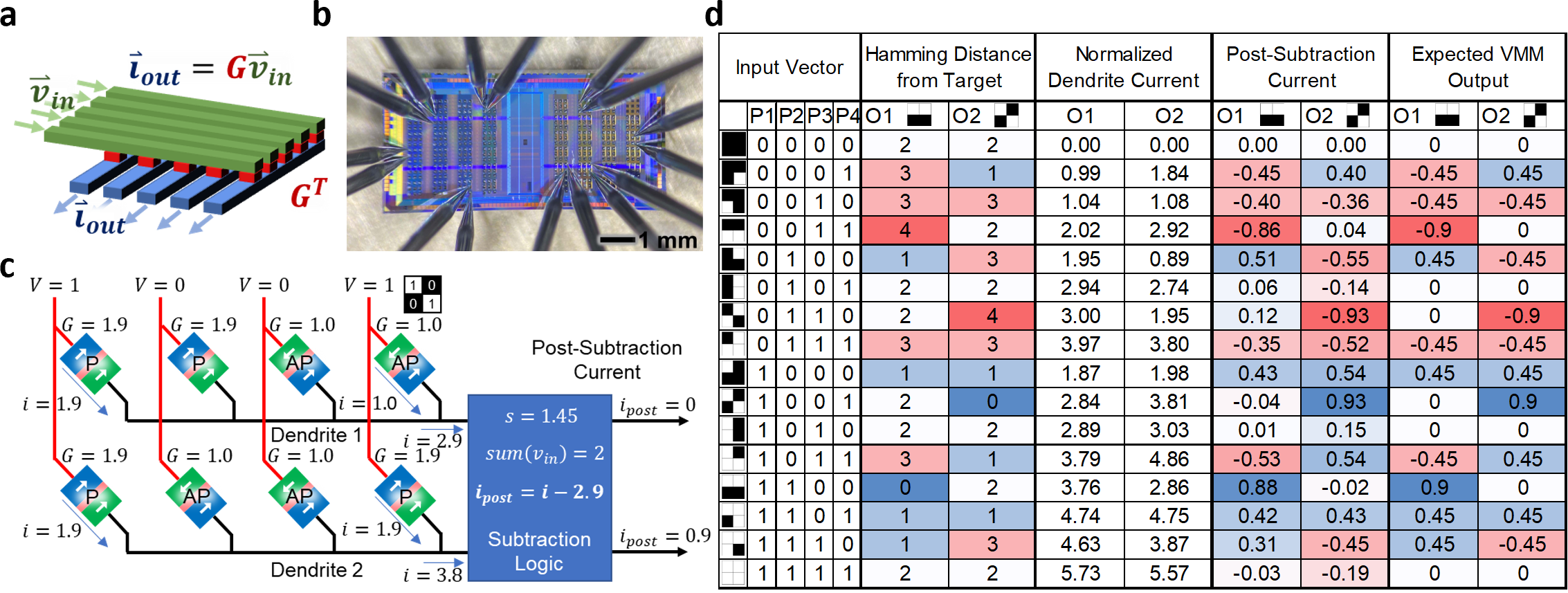}
    \caption{\textbf{MTJs for binary neuromorphic computing.} \textbf{a,} Vector-matrix multiplication (VMM) with a resistive memory crossbar array.  The matrix values are encoded in the device conductances, \textbf{\textit{G}}. The input vector is converted to an array of voltages fed into the column lines, $\vec{\textbf{\textit{v}}}_{in}$. When the row lines are grounded, the resulting current vector through them is the dendritic output of the VMM operation, $\vec{\textbf{\textit{i}}}_{out}$. This current can be sensed, and an analog-to-digital converter is used to convert it to digital signals. \textbf{b,} Eight MTJs from the depicted MTJ array were probed and connected. \textbf{c,} Schematic of the trained MTJ neuromorphic network.  Physical quantities are normalized such that  $v_{in} = 1/0$ and $G_{AP} = 1$.  Given this configuration, the output ranges between $I = \pm0.9$, with positive values corresponding to a greater match between the input image and the target. In this case, the input image (1001) is fed into the network matching target image 2 (1001) identically and target image 1 (1100) neutrally. \textbf{d,} All possible input images were presented to the MTJ VMM network.  The Hamming distances between the input and target images was calculated by the neuromorphic network. DC input voltages were fed into the network and the dendrite currents were measured and normalized by the average current through the AP MTJs. The subtraction factor was calculated to be 1.44 after normalization and was subtracted from the measured dendrite currents giving the VMM results shown. A color scale from blue to red is used to show image alignment with blue (red) indicating maximum (minimum) alignment, demonstrating successful VMM experimental calculation for all input combinations.}
    \label{fig:4x2experiment}
\end{figure}

We performed the first experimental demonstration of a neuromorphic network directly implemented with MTJ synapses, in accordance with the conventional neuromorphic VMM approach of Fig. \ref{fig:4x2experiment}a. With each MTJ conductance state representing a binary synaptic weight, a vector of input voltages applied to an MTJ synaptic array results in a dendritic output vector of currents. With eight in-plane STT-MTJ devices fabricated by Everspin (Fig. \ref{fig:4x2experiment}b), we implemented the 4x2 neuromorphic network of Fig. \ref{fig:4x2experiment}c.

In this experimental demonstration of neuromorphic inference, the network was tasked with determining the Hamming distance between input and target 2 pixel x 2 pixel images. The 4x2 STT-MTJ network was programmed to encode the two target images through a simple supervised approach, with black `0' pixels (white `1' pixels) encoding a synaptic weight of -0.45 (+0.45) and represented by the low-conductance AP (high-conductance P) state. Input images were applied to the network columns as binary voltages, with black `0' pixels (white `1' pixels) represented by a `0' (`1') voltage. Current therefore flowed through the network as a function of the input voltages and MTJ states, with the current measured for dendrite (row) outputs. To overcome the low on-off resistance ratio of MTJs, the number of input `1' voltages is subtracted from the dendrite current measurements through simple logic that enables negative weights with minimal overhead.

The neuromorphic network successfully determined the Hamming distance for all 16 2x2 pixel images, with the normalized post-subtraction currents showing high fidelity to the expected VMM in Fig. \ref{fig:4x2experiment}d. By setting proper thresholds, the post-subtraction results can be categorized into five distinct Hamming distance bins. As the variations from the expected output arose primarily due to power supply voltage variation and differences in probe connectivity, significantly decreased variation is expected for future neuromorphic networks with MTJ arrays directly integrated with the peripheral CMOS circuits. As shown in \cite{zhou2021experimental}, MTJ device-to-device variations do not impede the scaling of neuromorphic inference networks to large sizes.

\section*{Unsupervised Hebbian learning with STT-MTJ stochastic switching}

\begin{figure}[t!]
    \captionsetup{labelformat=simple}
    \centering
    \includegraphics[width=1\textwidth]{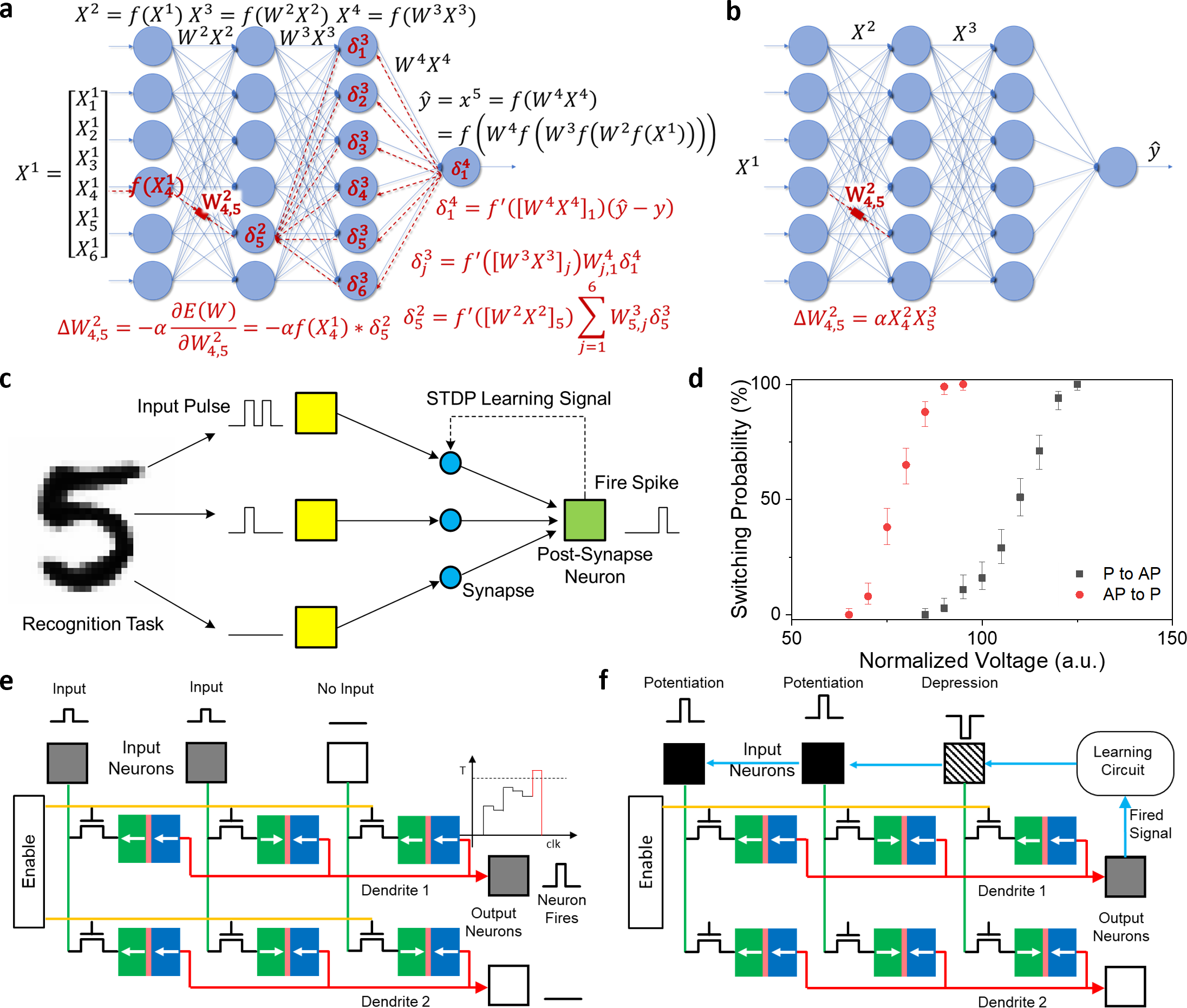}
    \caption{\textbf{Unsupervised learning with stochastic STT-MTJ switching.} \textbf{a-b,} Backpropagation vs. localized Hebbian learning. For both, an input vector $X^1$ is processed by the network via multiplication by weight matrices $W^i$ for layers $i$ to produce VMM output $\hat{y}$. \textbf{a,} For training via backpropagation, the impact of each weight on all downstream neurons must be calculated in order to minimize the error between $\hat{y}$ and the correct result ($y$). This is most efficiently performed by calculating derivatives from the downstream neurons and propagating these partial results ($\delta^i$) backward to each individual weight, which requires a large number of multiply-accumulate calculations for every classification performed by the network. \textbf{b,} For each synapse weight update using a localized Hebbian learning rule, a simple calculation is performed based on the behavior of the pre-synaptic and post-synaptic neurons. This localized learning requires significantly simpler computation, is highly parallelizable, more closely mimics synapse learning in the brain, and can be efficiently implemented in hardware. (Continued on next page)}
\end{figure}
\begin{figure}[t!]
    \captionsetup{labelformat=empty}
    \ContinuedFloat
    \centering
    \caption{(Continued from previous page) \textbf{c,} The SNN performs feed-forward recognition and feedback learning with an STDP learning rule. The input information is binarized to the input pulses, and the output information is represented by the output pulses from the output neuron. \textbf{d,} The stochastic switching probability measured for an STT-MTJ, where the different voltage levels with the same pulse durations write the MTJ to the parallel and anti-parallel states with different probabilities. \textbf{e,} Input neurons provide voltage pulses that flow through MTJ synapses and are integrated by the output neurons. The output neurons fire when their integration level is greater than the neuron threshold. \textbf{f,} During the learning process, a neuron firing is processed by a digital learning circuit to generate potentiation and depression pulses through the synapses connected to the fired neuron. These pulses probabilistically switch the MTJ synapses through stochastic STT-MTJ switching.}
    \label{fig:mram_snn}
\end{figure}

While binary STT-MTJ synapses are not prone to the deleterious impacts of memristor/PCM stochasticity on neuromorphic inference accuracy, the ability to leverage stochastic STT-MTJ synaptic switching for unsupervised Hebbian learning is of even greater import. Specifically, the ability for an artificial intelligence system to learn autonomously from unlabeled data after deployment could open a wide range of revolutionary new health, safety, military, and other applications. Supervised learning approaches, however, are limited by their requirement of labeled data and large energy costs, whereas unsupervised learning can be achieved with the unlabeled data available to autonomous systems.

As illustrated in Fig. \ref{fig:mram_snn}a, the backpropagation algorithm conventionally used for supervised learning continually updates synaptic weights to increase the matching between the network calculation and the correct output through numerous complex mathematical operations that are difficult to efficiently implement in a circuit. In contrast, Hebb postulated in 1949 \cite{hebb1949organization} that neurobiological learning is based on local activity, famously summarized as ``cells that fire together wire together,'' a far simpler process illustrated in Fig. \ref{fig:mram_snn}b. This Hebbian principle underlies the STDP learning rule depicted in Fig. \ref{fig:mram_snn}c, in which the relative timing between pre-synaptic and post-synaptic spikes governs synaptic weight updates. The simplicity and locality of the mathematical operations underlying this STDP learning process is well-suited to efficient hardware implementations of unsupervised Hebbian learning, as suggested by Querlioz \textit{et al.} for both analog memristor synapses \cite{querlioz2012bioinspired} and binary STT-MTJ synapses with stochastic switching \cite{vincent2015spin}.

We therefore propose a complete neuromorphic system that performs unsupervised Hebbian learning with a synchronous circuit that leverages the stochastic switching of STT-MTJ synapses to provide the analog behavior critical to learning while maintaining the stable binary conductance states necessary for high-accuracy inference. The application of a voltage across the MTJ causes the free layer to switch through STT, with a probability related to the voltage magnitude and pulse duration; Fig. \ref{fig:mram_snn}d shows the dependence of this stochastic switching probability on voltage magnitude. Whereas this stochastic STT switching is a challenge that must be overcome for MRAM, Querlioz \textit{et al.} showed that this stochasticity provides MTJs with the analog characteristics necessary for learning despite exhibiting only binary states \cite{vincent2015spin}. The stochastic nature of the STT-MTJ device switching enables analog learning at the system level, providing a best-of-both-worlds scenario that permits the simultaneous practicality and accuracy advantages of the binary STT-MTJ conductance states.

The operation of this Hebbian neuromorphic learning system is illustrated in the circuit diagrams of Fig. \ref{fig:mram_snn}e,f. Input data is provided as binary voltages, causing currents to flow through the MTJ synapses and dendrites to the output neurons, calculating the VMM as in Fig. \ref{fig:4x2experiment}. The dendrite currents are integrated by a standard CMOS circuit that emulates a leaky integrate-and-fire neuron, with the firing threshold continually updated by a homeostasis algorithm. When the integrated neuron signal is greater than the neuron's threshold, the output neuron generates a firing signal while adhering to a winner-take-all rule. This firing signal implements Hebb's principles through a simple STDP learning rule that applies learning pulses to each synapse connected to the output neuron that fired (\textit{i.e.,} in the same row); synapses connected to input neurons that previously fired within a specified number of clock cycles receive a potentiation pulse, whereas synapses connected to neurons without recent firing activity receive a depression pulse. The potentiation (depression) pulses cause the STT-MTJs to switch to the parallel (anti-parallel) states with a particular probability, based on the binarized stochastic learning rule \cite{vincent2015spin}. The proposed unsupervised Hebbian learning system thus leverages stochastic STT-MTJ switching in a manner that facilitates fabrication through standard foundry processes for digital CMOS circuits and MTJ crossbar arrays.

\section*{Experimental demonstration of unsupervised Hebbian learning network with STT-MTJ synapses}

We experimentally demonstrated this unsupervised Hebbian learning with a 4x2 neuromorphic network of STT-MTJ synapses, as shown in Fig. \ref{fig:4x2_learning}. Based on the probe station setup that was used for the STT-MTJ neuromorphic inference network, we developed a platform for demonstrating unsupervised Hebbian learning with a manual control switch array. We used this switch array to manually implement the neuron behavior through the appropriate connections of the MTJ fixed layers to the inference (V$_\mathrm{I}$), potentiation (V$_\mathrm{P}$), and depression (V$_\mathrm{D}$) voltages; this switch array was also used to connect the dendrites to the SMUs and ground, as appropriate. For this experiment, the STDP switching probabilities were set to 35\% for AP to P and 30\% for P to AP through control of the pulse duration and voltages. This experiment is thus a complete and direct demonstration of the system described in the previous section and Fig. \ref{fig:mram_snn}, with all of the non-MTJ components implemented through manual control of the switch array.

\begin{figure}[t!]
    \captionsetup{labelformat=simple}
    \centering
    \includegraphics[width=1\textwidth]{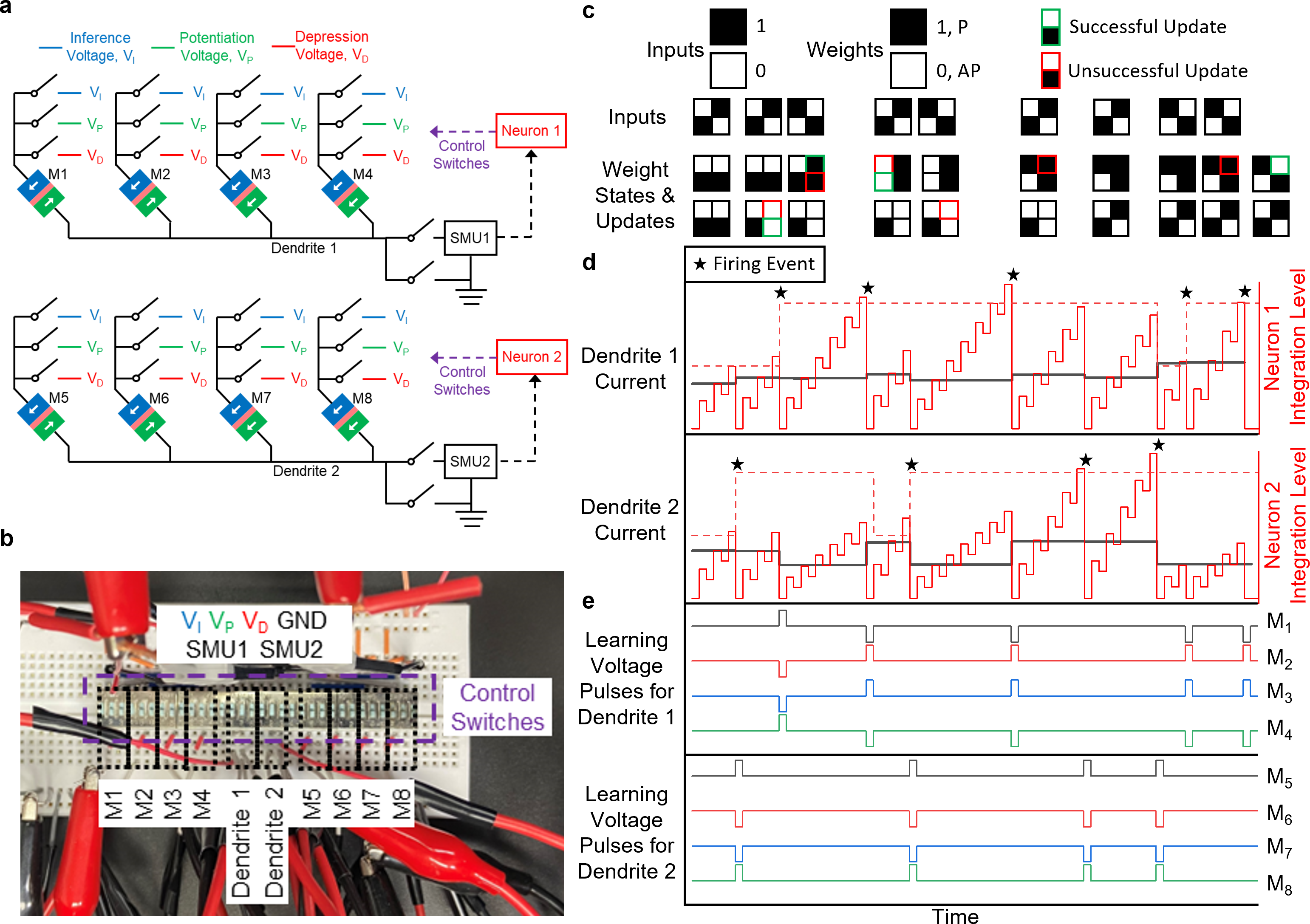}
    \caption{\textbf{Experimental demonstration of unsupervised Hebbian learning with stochastic STT-MTJ switching.} \textbf{a,} Schematic of the 4x2 neuromorphic learning network, with read (V$_\mathrm{I}$), potentiation (V$_\mathrm{P}$), and depression (V$_\mathrm{D}$) voltages connected to each STT-MTJ synapse (M1-M8) by manual switches. The V$_\mathrm{I}$ switch is closed for input `1' signals, and left open for input `0' signals. Manual switches connect the dendrites to source measure units (SMUs) during inference clock cycles and to ground during learning clock cycles, in accordance with the LIF neuron function. \textbf{b,} Experimental platform including manual control switch array. M1-M8 are connected to the fixed layers of the MTJs (blue in \textbf{a}), while dendrites 1 and 2 are connected to the free layers (green in \textbf{a}) of M1-M4 and M5-M8, respectively. \textbf{c,} Unsupervised Hebbian learning task inputs and experimental outputs. Two 2x2 pixel input images are repeatedly presented to the neuromorphic network (1001 and 0110), in random order. After each image is presented, the network learns by applying the unsupervised Hebbian learning rule to update the states of the MTJ synapses. This state updating proceeds probabilistically, according to the stochastic STT switching probabilities for the potentiation and depression pulses. Whereas the synapses connected to both neurons are initially in the same states (1100), this stochastic Hebbian learning successfully causes neuron 1 to recognize one input image (0110) while neuron 2 recognizes the other input image (1001). (Continued on next page)}
\end{figure}
\begin{figure}[t!]
    \captionsetup{labelformat=empty}
    \ContinuedFloat
    \centering
    \caption{(Continued from previous page) \textbf{d,} Electrical measurements and neuron integration. For each input image, the dendrite currents (solid black line) were measured with the SMUs (similar to the VMM inference measurements of Fig. 1). The dendrite current signals were integrated and leaked via the neuron firing algorithm until one of the neuron integration levels (solid red line) surpassed its threshold (dashed red line), at which point that neuron fired. Following each firing event, the neuron integration levels were set to zero and the neuron thresholds were adjusted according to the homeostasis algorithm. \textbf{e,} STDP learning voltage pulses. Following each firing event, stochastic STT learning pulses were provided to the MTJs connected to the fired neuron. MTJs that had received a  `1' input are provided a positive potentiation learning voltage, while MTJs that had received a `0' input are provided a negative depression learning voltage. As STT switching of MTJs is stochastic, truly random behavior determined whether the MTJ switching attempts were successful (green and red squares in \textbf{c}). When a potentiation (depression) pulse is provided to an MTJ that is already in the parallel (anti-parallel) state, no switching is attempted.}
    \label{fig:4x2_learning}
\end{figure}

To observe Hebbian learning with this neuromorphic STT-MTJ network, we challenged the network with a simple image clustering task with two images: one output neuron must learn to recognize one input image while the other output neuron learns to recognize the other output image. As shown in Fig. \ref{fig:4x2_learning}c, the STT-MTJs connected to both neurons were initialized to the same state with two parallel and two anti-parallel MTJs. A series of input images was provided to the network, with each input image applied until one of the two output neurons fired (as shown in Fig. \ref{fig:4x2_learning}d). Whenever a neuron fired, STDP learning pulses were applied to all of the STT-MTJs connected to the neuron that fired: MTJs that had received a  `1' (`0') input were provided a potentiation (depression) pulse (Fig. \ref{fig:4x2_learning}e).

In response to these Hebbian learning pulses, the stochastic nature of STT-MTJ switching caused one of the following to occur:
\begin{itemize}
    \item If the learning pulse was directed toward the state that the MTJ was already in, no switching occurred. That is, no change in state was caused by the application of a potentiation (depression) pulse to an MTJ that was already in the parallel (anti-parallel) state.
    \item As a result of the random behavior of the stochastic STT-MTJ switching, the MTJ successfully switched.
    \item As a result of the random behavior of the stochastic STT-MTJ switching, the MTJ did not switch.
    \end{itemize}
As can be seen in Fig. \ref{fig:4x2_learning}c, there are six red squares and four green squares, indicating that the STT-MTJ stochastic switching had a 40\% switching rate within this small sample. After the application of nine input images and the corresponding STDP learning pulses, the STT-MTJ synaptic weights completed the learning process and became identical to the two input images, with one output neuron having learnt to recognize one input image while the other output neuron learnt to recognize the other output image. This unsupervised Hebbian learning was thus successful in causing the neuromorphic network to learn to perform this clustering task.

\section*{Unsupervised Hebbian learning and recognition of MNIST handwritten digits}

To demonstrate the ability of the proposed unsupervised Hebbian learning system to scale to large neuromorphic networks of STT-MTJs, system-level simulations were performed in C++ on a binarized MNIST handwritten digit dataset \cite{lecun1998gradient}. The binarized 28x28 input dataset was provided to the 784 input neurons, and varying quantities of output neurons were used. In some network configurations, several synapses in the same row can connect to the same input neurons to compensate for the STT-MTJ synapse stochasticity; this quantity is referred to as \textit{r} (values of 1, 2, 4, and 8 were considered), and these synapses all receive the same STDP learning pulses.

\begin{figure*}[t!]
    \centering
    \includegraphics[width=1\textwidth]{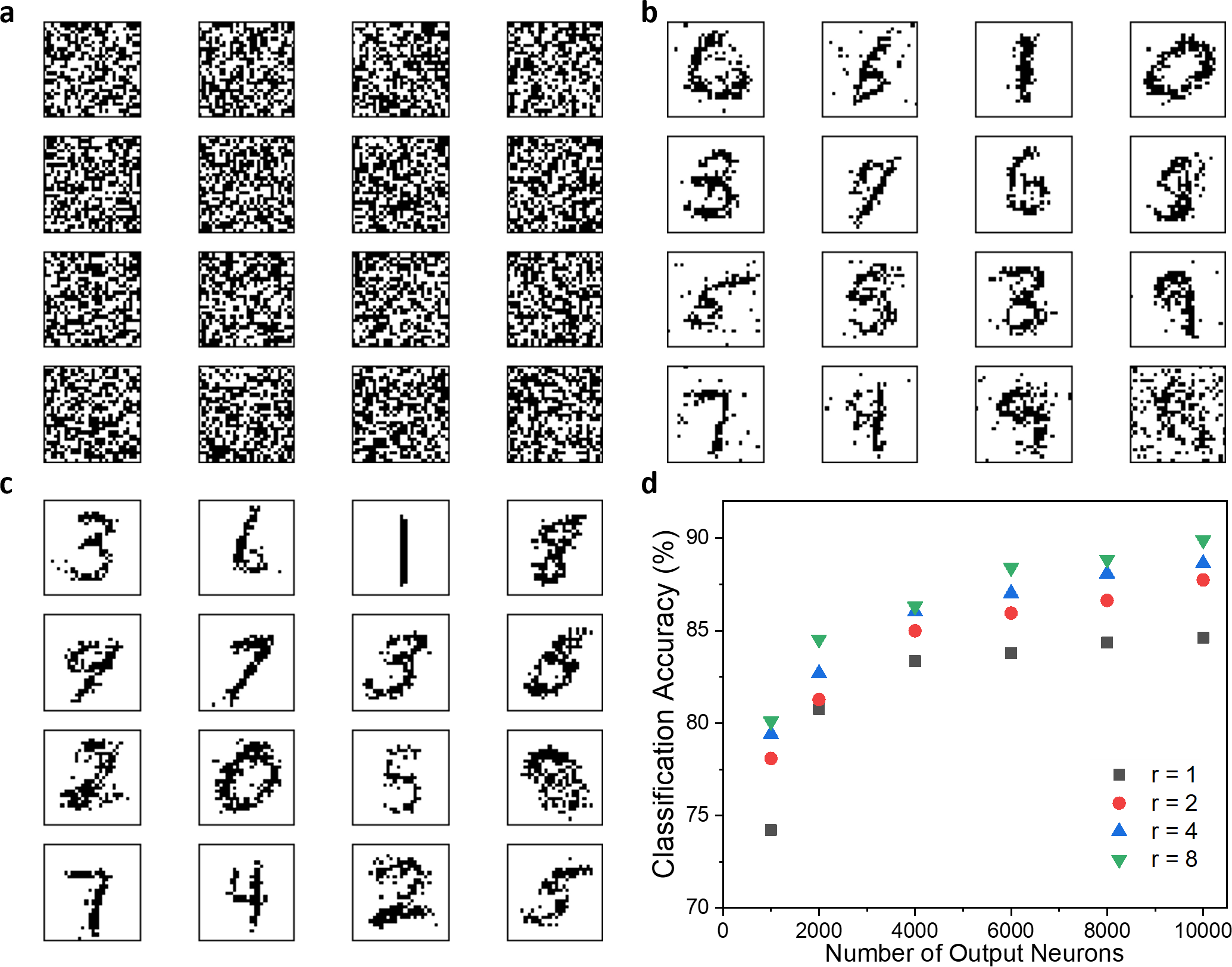}
    \caption{\textbf{Behavioral simulation of unsupervised Hebbian learning of the MNIST handwritten digit dataset with stochastic STT-MTJ switching.} \textbf{a,} All the STT-MTJ synapses are initialized to random states before learning. \textbf{b,} The synapse states of selected neurons after training a 500-neuron network with 10,000 MNIST handwritten digits. \textbf{c,} The states of the same neurons after training the network with 60,000 MNIST handwritten digits. Note that several neurons have specialized in different digits than in \textbf{b}. \textbf{d,} Accuracy as a function of the number of output neurons, where \textit{r} is the number of STT-MTJ synapses representing each pixel.}
    \label{fig:simu_result}
\end{figure*}

The simulation methodology is identical to the experiment of Fig. \ref{fig:4x2_learning}. First, all synapses are initialized to random states. After the training images are binarized by comparing the gray-scale pixel data to a threshold, the output neurons are labeled based on their firing activity. The recognition accuracy is then calculated based on the percentage of the 10,000 testing images that are correctly labeled by the neurons. Fig. \ref{fig:simu_result}a-c show the evolution of MTJ synapse weights during the simulation as a result of the STDP learning rule. As can be seen in the figure, some neurons begin to specialize in one digit but later evolve their specialization to another digit in response to the firing activity of other output neurons.

The inference accuracy results for a single layer STT-MTJ SNN are shown in Fig. \ref{fig:simu_result}d for varying quantities of output neurons and synapses for each pixel. The inference accuracy clearly increases with both increasing numbers of output neurons and increasing numbers of synapses sharing each pixel. When there are 10,000 output neurons and eight synapses for each pixel, the inference accuracy can reach 90\%. These results are comparable to simulations of unsupervised single-layer SNNs based on multilevel memristor evaluated with a similar size and methodology \cite{rathi2018stdp}, demonstrating the feasibility of the proposed binary STT-MTJ approach. 

However, multilevel and analog behavioral simulations such as \cite{rathi2018stdp} do not consider the challenges for precisely writing and storing memristor resistance states. The analog and multilevel states of memristor and PCM devices exhibit stochastic writing, weight drift over time, and limited endurance that reduces the precision of synapse weights, leading to drastic decreases in system accuracy as thoroughly shown in \cite{xiao2021accuracy}. Given that the proposed use of STT-MTJs achieves similar inference accuracies as memristors/PCM without making unreasonable assumptions about STT-MTJ behavior, the proposed binary MTJs with stochastic STT switching have the potential to provide higher accuracies than can be achieved with memristors and PCM.

\section*{Conclusions}

This work presents the first experimental demonstrations of neuromorphic networks directly implemented with MTJ synapses for both inference and learning, while also proposing a synchronous SNN system that leverages the stochastic switching of STT-MTJs for unsupervised Hebbian learning. 4x2 neuromorphic networks of STT-MTJs are used to perform image recognition on 16 2x2 pixel images pre-trained in a supervised manner, as well as unsupervised clustering of two 2x2 pixel images through stochastic STDP learning. Simulation results of unsupervised Hebbian learning on a large-scale neuromorphic network achieve 90\% inference accuracy on the MNIST handwritten digit recognition dataset, indicating the ability to achieve high accuracy for commercially-relevant artificial intelligence applications. By exploiting the analog nature of stochastic STT switching in concert with the stable binary MTJ states, the proposed neuromorphic SNN takes full advantage of the STT-MTJ physics to provide the best of both worlds for artificial intelligence hardware.

\bibliographystyle{naturemag}
\bibliography{ref} 

\bigskip
\large
\noindent\textbf{Acknowledgments}
\normalsize

\noindent This work was supported in part by Semiconductor Research Corporation (SRC) Task No. 2810.030 through UT Dallas’ Texas Analog Center of Excellence (TxACE) and the National Science Foundation under CCF Award No. 2146439.

\end{document}